\documentclass[runningheads]{llncs}
\usepackage[T1]{fontenc}
\usepackage{graphicx}
\usepackage{amsmath}
\usepackage{amsfonts}
\usepackage{multirow}
\usepackage[misc]{ifsym}
\begin{document}
\title{Dynamical Graph Echo State Networks with Snapshot Merging for Dissemination Process Classification}
\titlerunning{The GDGESN with snapshot merging for DPC tasks}

%
%
\author{Ziqiang Li\Letter\inst{1}\orcidID{0000-0002-7208-9003} \and
Kantaro Fujiwara\inst{1}\orcidID{0000-0001-8114-7837}
\and
Gouhei Tanaka\inst{1,2}\orcidID{0000-0002-6223-4406}}
\authorrunning{Z. Li et al.}
%
\institute{International Research Center for Neurointelligence, The University of Tokyo, Tokyo 113-0033, Japan \\
\email{\{ziqiang-li,kantaro\}@g.ecc.u-tokyo.ac.jp}\\
\and
Department of Computer Science, Graduate School of Engineering, Nagoya Institute of Technology, Nagoya 466-8555, Japan \\
\email{gtanaka@nitech.ac.jp}
}
\maketitle              
\begin{abstract}
The Dissemination Process Classification (DPC) is a popular application of temporal graph classification. The aim of DPC is to classify different spreading patterns of information or pestilence within a community represented by discrete-time temporal graphs. Recently, a reservoir computing-based model named Dynamical Graph Echo State Network (DynGESN) has been proposed for processing temporal graphs with relatively high effectiveness and low computational costs. In this study, we propose a novel model which combines a novel data augmentation strategy called snapshot merging with the DynGESN for dealing with DPC tasks. In our model, the snapshot merging strategy is designed for forming new snapshots by merging neighboring snapshots over time, and then multiple reservoir encoders are set for capturing spatiotemporal features from merged snapshots. After those, the logistic regression is adopted for decoding the sum-pooled embeddings into the classification results. Experimental results on six benchmark DPC datasets show that our proposed model has better classification performances than the DynGESN and several kernel-based models. 

\end{abstract}
\section{Introduction}
\label{Sec:Introduction}
The dissemination process is used to describe the spreading of information (e.g. fake news and rumors) or infectious diseases (e.g. Covid-19 and meningitis) within a community. Since dissemination patterns of virus strains or different rumors are various, it is hard to recognize them within a relatively short period of time accurately. Based on this background, Dissemination Process Classification (DPC) is a highly-demanded technology for experts in relevant fields to distinguish them before carrying out possible interventions and countermeasures. 
\par
\begin{figure}[t]
    \centering
    \includegraphics[scale=0.4]{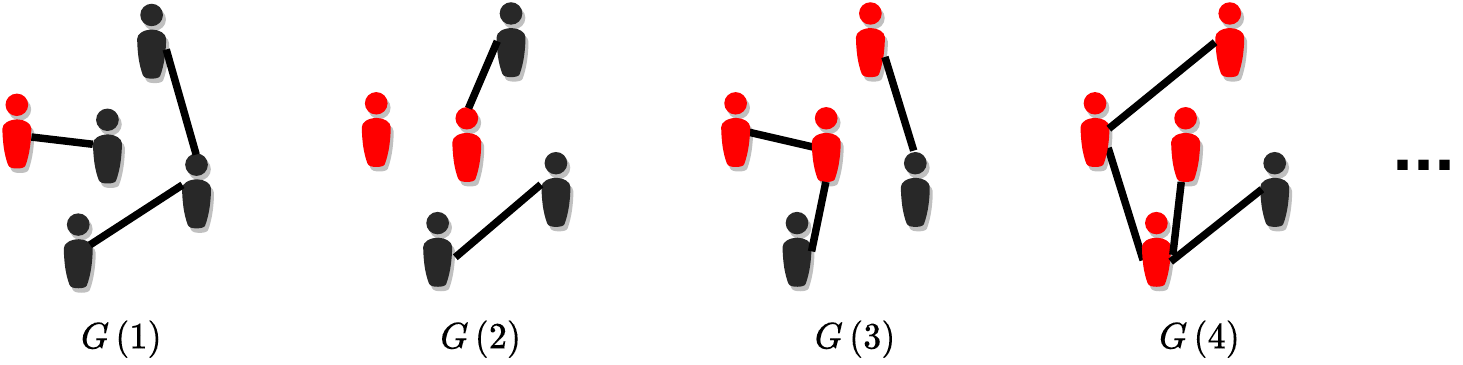}
    \caption{An example of epidemic spreading. Icons marked in red and black represent infected and uninfected people, respectively, and a black line between two icons indicates that contact exists between two people at that time step.}
    \label{fig:example_dissemination}
\end{figure}

Normally, dissemination processes can be represented by temporal graphs with dynamic connections and temporal signals. We show an example of an epidemic spreading in Fig~\ref{fig:example_dissemination}, where $G(t)$ means the $t$-th snapshot of the temporal graph $\mathcal{G}$. We can notice that uninfected people (marked in black) can be infected probabilistically by contact with infected people (marked in red). Usually, one kind of epidemic has its own basic reproduction number, which leads to different dissemination processes.
\par
Generally, DPC can be turned into a Discrete-time Temporal Graph (DTG) classification task. To deal with this task, advanced deep learning models ~\cite{chen2018gc,guo2019attention,li2021spatial} designed by combining variants of Graph Convolutional Networks (GCNs)~\cite{kipf2016semi} with those of Recurrent Neural Networks (RNNs)~\cite{cho2014properties,hochreiter1997long} and/or the attention mechanism~\cite{vaswani2017attention} are widely considered to be ideal choices. The common ground of these models in structure is leveraging multiple graph convolution layers for extracting spatial features and using recurrent layers or attention layers for mining the temporal relationships. In this regard, high computational costs need to be spent to obtain a well-trained complex model. Furthermore, extra efforts for solving gradient explosion and vanishing problems are unavoidable. Another direction is to use some transformation methods to stitch the sequential snapshots into a large-scale static graph and then apply some graph kernel methods (i.e. the Weisfeiler-lehman graph kernels~\cite{shervashidze2011weisfeiler}) to generate the final classification results. Methods following this direction can avoid effects on capturing the temporal dependency in the snapshot sequence, but the large-scale static graphs lead to significantly high computational costs in the calculations of the gram matrix for a support vector machine~\cite{hearst1998support} by using those graph kernels~\cite{oettershagen2020temporal}.
\par
Reservoir Computing (RC)~\cite{lukovsevivcius2012practical,tanaka2019recent} is an efficient framework derived from RNNs, which maps the sequential inputs into high dimensional spaces through a predetermined dynamical system. This characteristic enables the training costs of its derived models to be remarkably lower than those of fully-trained RNNs. The Echo State Network (ESN)~\cite{jaeger2002tutorial}, as one of the representative models of RC, and its variants have been intensively studied for handling various time series processing tasks~\cite{gallicchio2017deep,li2022multi}. Recently,  D. Tortorella \& A. Micheli successfully extended the standard ESN to a novel RC model called Dynamical Graph Echo State Network (DynGESN)~\cite{tortorella2021dynamic}, which is capable of dealing with discrete-time temporal graphs processing tasks. 
\par 
A recent work has demonstrated that DynGESN outperforms some kernel-based methods on a number of DPC benchmark datasets~\cite{micheli2022discrete}. However, we noticed that only one single dynamical characteristic included in the original temporal graphs is extracted in DynGESN~\cite{micheli2022discrete}. Obviously, this monotonous strategy may hinder the model from extracting diverse dynamical characteristics extended from the original temporal snapshots.
\par
To solve this problem, a new model, Grouped Dynamical Graph Echo State Network (GDGESN), is proposed for DPC tasks in this study. This model can extract various spatiotemporal features from augmented inputs by group-wise reservoir encoders and generate accurate dissemination classification results by a linear classifier efficiently. In this regard, we propose a simple augmentation strategy called snapshot merging to generate multi-timescale temporal graphs and then leverage the multiple-reservoir framework~\cite{li2023multi} to build the group-wise reservoir encoders. We execute experiments for comparing the classification performances with those of the DynGESN and some kernel-based methods on six benchmark DPC datasets. The experimental results show that the accuracies of our model are higher than those of DynGESN and are close to those of kernel-based methods on some DPC datasets, which manifests the GDGESN owns relatively high effectiveness in dealing with DPC tasks.
\par
The rest of this paper is organized as follows: The preliminary about temporal graphs is introduced in Section~\ref{Sub:Preliminaries}. The proposed method is described in Section~\ref{Sec:proposed model}. The analysis about the computational complexity of the proposed model is presented in Section~\ref{Sec:computational complexity}. The details of the experiments are introduced in Section~\ref{Sec:Experiments}. The discussion is given in Section~\ref{Sec:Conclusion}.

\section{Preliminaries}
\label{Sub:Preliminaries}
Generally, a discrete-time temporal graph is composed of a sequence of snapshots, which can be denoted by $\mathcal{G}=\left \{ G\left ( t \right )  \right \} _{t=1}^{N_{T}}$, where $G\left ( t \right )$ is the snapshot at time $t$ and $N_{T}$ is the length of $\mathcal{G}$. The snapshot $G\left ( t \right ) = \left\{ \mathbf{v}\left ( t \right ), \mathbf{A}\left ( t \right ) \right\}$ contains a time-varying vertex signal vector $\mathbf{v}\left ( t \right )\in \mathbb{R}^{N_{V}}$ and the corresponding adjacency matrix $\mathbf{A}\left ( t \right )\in \mathbb{R}^{N_{V}\times N_{V}}$, where $N_{V}$ is the number of vertices. The state value of the $i$-th vertex at time $t$ can be represented by $v_{i}(t)\in \mathbb{R}$. For representing the dissemination process, we define that $v_{i}(t) =1$ if the $i$-th person is affected at time $t$ and $v_{i}(t) =0$ otherwise. We suppose that each graph is undirected, which can be represented by $A_{i,j}\left ( t \right )  = A_{j,i}\left ( t \right ) = 1$ if there is a contact between the $i$-th person and the $j$-th person, $A_{i,j}\left ( t \right )  = A_{j,i}\left ( t \right ) = 0$ otherwise. Moreover, we assume that a dissemination process classification dataset has $N_{S}$ temporal graphs and the corresponding labels, which can be represented by $\left \{ \mathcal{G}_{s}, \mathbf{y}_{s} \right \} _{s=1}^{N_{s}}$, where $\mathbf{y}_{s}\in \mathbb{R}^{N_{Y}}$ is the label represented by the one-hot encoding for $\mathcal{G}_{s}$.
\section{The proposed model}
\label{Sec:proposed model}
\begin{figure}
    \centering
    \includegraphics[scale=0.35]{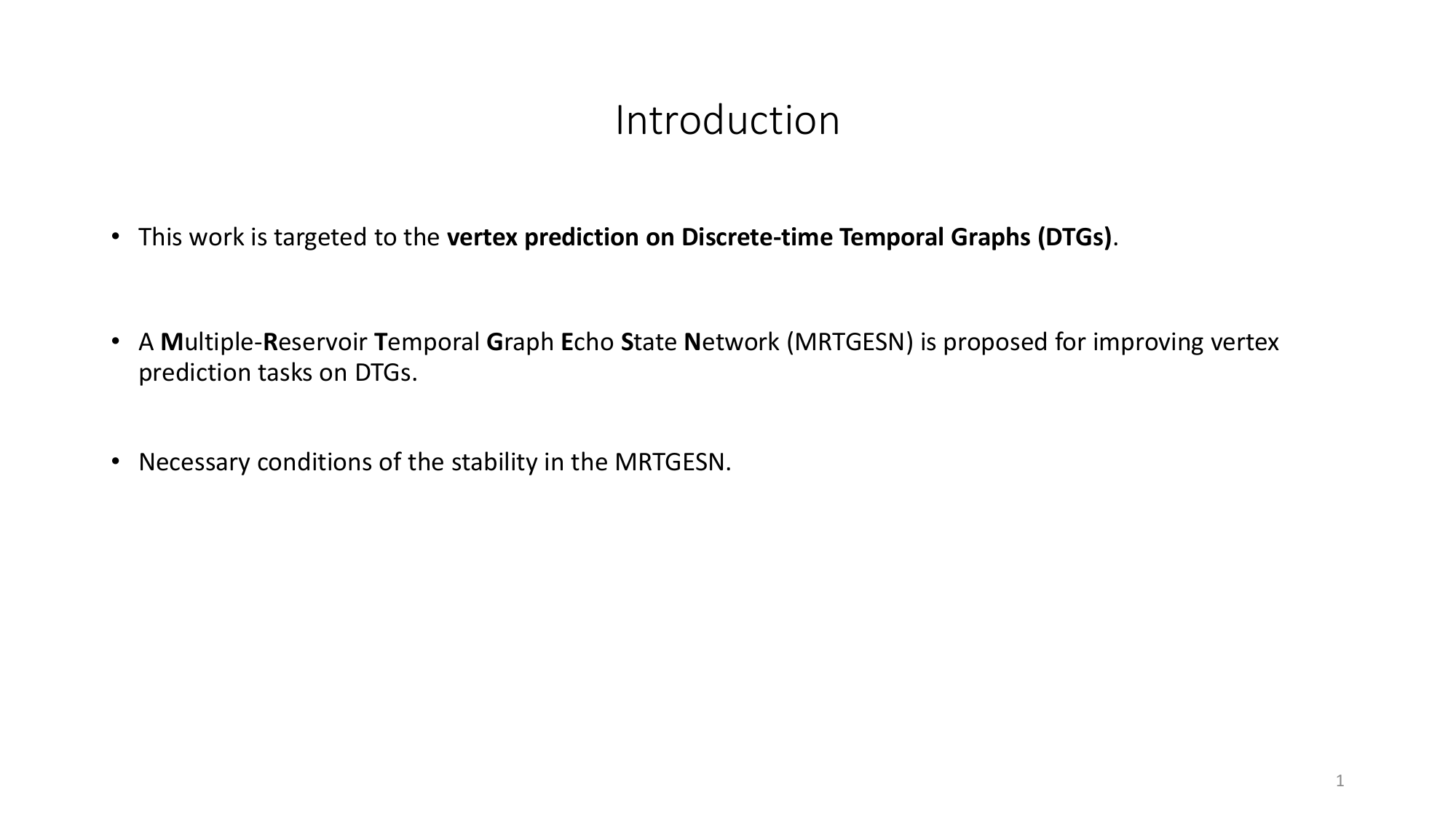}
    \caption{An example of the proposed model with three groups of reservoir encoders.}
    \label{fig:GDGESN}
\end{figure}
A schematic diagram of the GDGESN is shown in Fig.~\ref{fig:GDGESN}. This is a case where a dissemination process represented by a discrete-time temporal graph is fed into the GDGESN with three groups of reservoir encoders. We can notice that the model consists of three components, including a merged snapshot converter, a set of multiple-reservoir encoders, and a linear classifier. In the merged snapshot converter, a DTG is transformed into three new DTGs with different window sizes. In the multiple-reservoir encoder, each transformed temporal graph is fed into the corresponding group-wise reservoir encoders for generating various vertex embeddings. In the linear decoder, the aggregated embeddings of the last time step obtained by the sum-pooling operation are collected from all reservoir encoders and then decoded into the classified results. The details about the above-mentioned components are introduced in Sections~\ref{Subsec:The Merged snapshot converter},~\ref{Subsec:Multiple-reservoir encoders}, and \ref{Subsec:The linear classifier}, respectively.
\subsection{The merged snapshot converter}
\label{Subsec:The Merged snapshot converter}
\begin{figure}[t]
    \centering
    \includegraphics[scale=0.4]{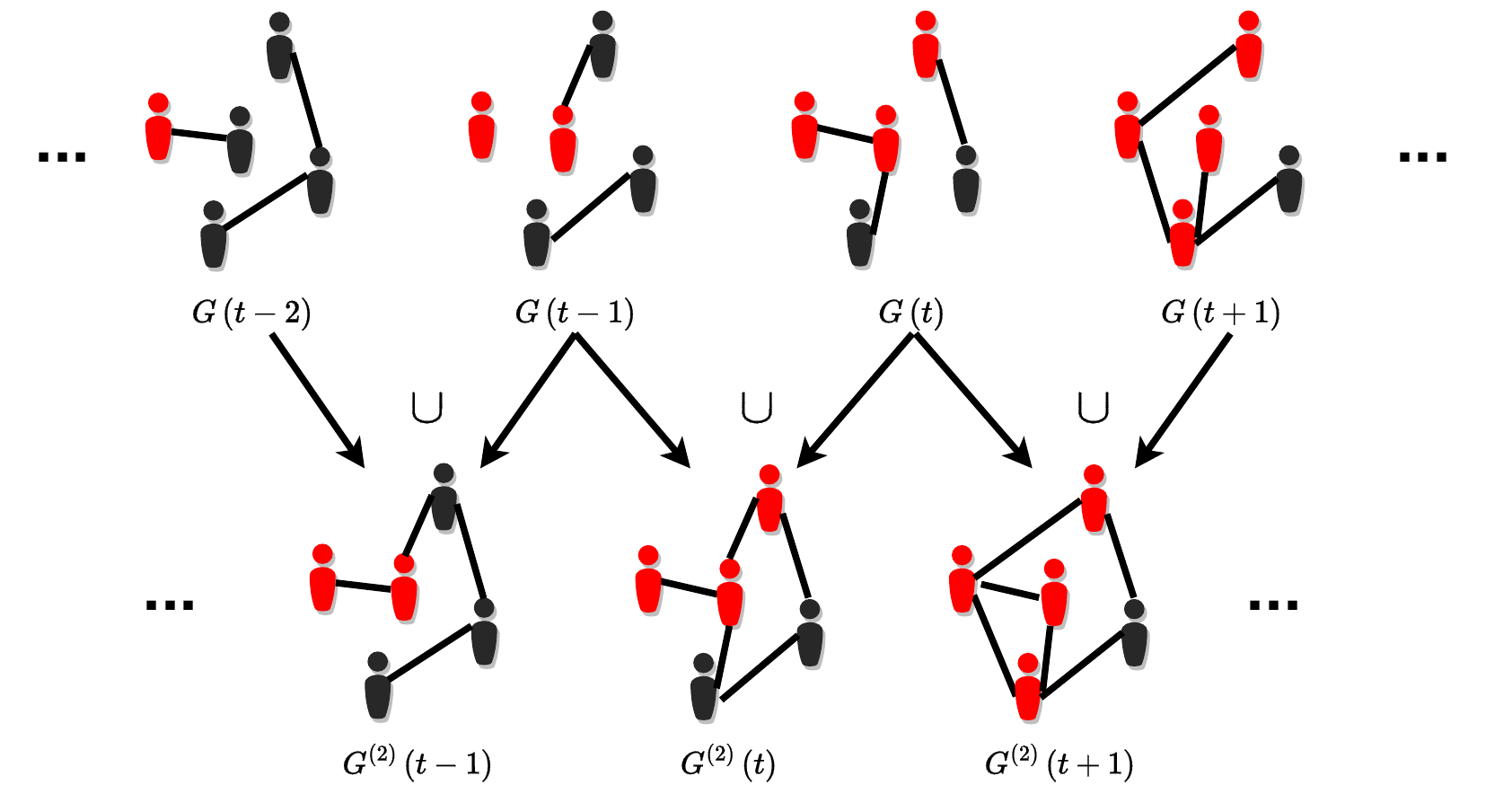}
    \caption{An example of transforming the original snapshot sequence into a merged snapshot sequence with the size of sliding window $\omega=2$.}
    \label{fig:example_sm}
\end{figure}

The merged snapshot converter is proposed to merge several neighboring snapshots into one merged snapshot. To this end, we define a window that slides on the zero-padded snapshot sequence. We denote the size of the sliding window by $\omega$. For simplicity, we fix the stride of this sliding window to be one. In order to keep the length of the merged snapshot sequence the same as that of the original snapshot sequence, we add $(\omega-1)$ empty snapshots into the beginning of the original snapshot sequence, which can be formulated as follows: 
\begin{equation}
\label{eq1}
    \mathcal{P}_{s} = \left \{ \underbrace{G_{nil},\ldots,G_{nil}}_{\omega-1},G\left ( 1 \right ),G\left ( 2\right ),\ldots,G\left ( N_{T}\right ) \right \},
\end{equation}
where $\mathcal{P}_{s}$ means the $s$-th snapshot-padded sequence with length $(N_{T}+\omega-1)$ and $G_{nil}$ represents the empty snapshot whose signal value of each vertex is zero. We assume that the 
merged temporal graphs corresponding to $N_{G}$ different sizes of the sliding windows can be organized into $N_{G}$ groups. Therefore, we represent the size of the sliding window corresponding to the $g$-th group as $\omega^{(g)}$ for $g = 1, 2, \dots, N_{G}$. Based on the above settings, We can merge snapshots into a new snapshot by executing the logical OR operation within a sliding window with size $\omega^{(g)}$, which can be formulated as follows:
\begin{equation}
\label{eq2}
\begin{matrix}
v_{i}^{\left ( g \right ) }\left ( t \right ) = v_{i}\left ( t-\omega^{\left ( g \right ) }+1 \right )  \cup v_{i}\left ( t-\omega^{\left ( g \right ) }+2 \right ) \cup \dots \cup v_{i}\left ( t \right ),
\\
A_{i,j}^{\left ( g \right ) }\left ( t \right ) = A_{i,j}\left ( t-\omega^{\left ( g \right )}+1 \right ) \cup A_{i,j}\left ( t-\omega^{\left ( g \right )}+2 \right )
\cup\dots \cup A_{i,j}\left ( t \right ).
\end{matrix}
\end{equation}
From Eq.~\ref{eq2}, we can obtain diverse spatiotemporal information with different $\omega^{\left ( g \right ) }$. Figure~\ref{fig:example_sm} shows an example of transforming the original snapshot sequence into a merged snapshot sequence by the merged snapshot converter with $\omega=2$. This example indicates that the merged snapshot converter can produce multi-timescale spatiotemporal inputs with different sizes of sliding windows. The experimental results presented in Section~\ref{subsec:Experimental results} demonstrate that the merged snapshot converter with various sizes of sliding windows can improve the classification performances of the GDGESN on some DPC datasets. 
\subsection{The multiple-reservoir encoder}
\label{Subsec:Multiple-reservoir encoders}
The multiple-reservoir encoder is proposed for extracting spatiotemporal features from merged snapshot sequences. We organize reservoir encoders following the  layout described in Ref.~\cite{li2022multi}. Note that a reservoir encoder denoted by $\Theta_{enc}$ contains an input weight matrix $\mathbf{W}_{in}\in \mathbb{R}^{N_{R}\times N_{U}}$ and a reservoir matrix $\mathbf{W}_{res} \in \mathbb{R}^{N_{R} \times N_{R}}$, where $N_{R}$ is the size of the reservoir. We add a superscript $\left ( g,l \right )$ to $\Theta_{enc}$ for indicating the encoder located at the $l$-th layer of the $g$-th group, which can be formulated by $\Theta_{enc}^{(g,l)} = \left \{ \mathbf{W}_{in}^{(g,l)}, \mathbf{W}_{res}^{(g,l)} \right \} $ for $1\le g\le N_{G}$ and $1\le l\le N_{L}$, where $N_{G}$ and $N_{L}$ are maximal numbers of groups and layers, respectively.
\par
In the encoding process, the vertex embedding matrix at time $t$, $\mathbf{X}^{\left ( g,l \right )}\left (t\right )\in \mathbb{R}^{N_{R}\times N_{V}}$, can be calculated as follows:
\begin{align}
\label{eq4}
\mathbf{X}^{\left ( g,l \right )}_{s}\left (t\right )  &= \alpha f\left ( \mathbf{W}_{in}^{\left ( g,l \right ) }\mathbf{U}^{\left ( g,l \right )}_{s}\left ( t \right )+\mathbf{W}_{res}^{\left ( g,l \right ) }\mathbf{X}^{\left ( g,l \right )}_{s}\left (t-1\right ) \mathbf{A}\left ( t \right )   \right ) \nonumber \\
&+ \left ( 1-\alpha  \right ) \mathbf{X}^{\left ( g,l \right )}_{s}\left (t-1\right ),
\end{align}
where $\alpha \in \left ( 0,1 \right ]$ is the leaking rate, $f\left ( \cdot \right )$ is an activation function, and $\mathbf{U}^{\left ( g,l \right )}$ is the input matrix used for receiving the various vertex inputs, i.e.
\begin{equation}
    \mathbf{U}^{\left ( g,l \right ) }_{s}\left (t\right )=\begin{cases}\mathbf{v}^{\left ( g \right ) }_{s}\left (t\right )
  & \text{ for } l= 1\\\mathbf{X}^{\left ( g,l-1 \right ) }_{s}\left (t\right )
  & \text{ for } l>1
\end{cases},
\end{equation}
the element values of $\mathbf{W}_{in}\in\mathbb{R}^{N_{R}\times N_{V}}$ are randomly chosen from a uniform distribution with the range of $\left [ -\eta,\eta \right ]$. The element values of $\mathbf{W}_{res}^{\left ( g,l \right ) }$ are randomly assigned from the uniform distribution $\left [ -1, 1 \right ]$. In order to ensure the echo state property~\cite{tortorella2021dynamic} in each encoder, we keep $\rho \left ( \mathbf{W}^{\left ( g,l \right ) }_{res} \right ) <1/ \rho \left ( \mathbf{A}_{s}\left (  t\right )  \right )$ at each time step.
\subsection{The linear classifier}
\label{Subsec:The linear classifier}
The recognition score of the $s$-th temporal graph, $\mathbf{\hat{y}}_{s}\in \mathbb{R}^{N_{Y}}$, is calculated through a simple linear mapping, which can be formulated as follows:
\begin{equation}
    \mathbf{\hat{y}}_{s} = \mathbf{W}_{out}\mathbf{c}_{s}+\mathbf{b},
\end{equation}
where $\mathbf{c}_{s}$ is the sum-pooled vector which can be calculated as follows:
\begin{equation}
   \mathbf{c}_{s}= \left [ sp \left ( \mathbf{X}_{s}^{\left ( 1,1 \right ) }(N_{T}) \right ) ; sp\left ( \mathbf{X}_{s}^{\left ( 1,2 \right ) }(N_{T}) \right ) ;\ldots; sp\left ( \mathbf{X}_{s}^{\left ( N_{G},N_{L} \right ) }(N_{T}) \right ) \right ] \in \mathbb{R}^{N_{R}N_{G}N_{L}},
\end{equation}
where $\left [ \cdot;\cdot \right ]$ represents the vertical concatenation and $sp\left ( \cdot  \right )$ acts for the operation of summing $N_{V}$ column vectors of $\mathbf{X}_{s}^{\left ( g,l \right ) }\left ( N_{T} \right ) $ up. The readout matrix $\mathbf{W}_{out}\in \mathbb{R}^{N_{Y}\times N_{R}N_{G}N_{L}}$ can be calculated as follows:
\begin{equation}
    \mathbf{W}_{out} = \mathbf{Y}\mathbf{C}^\mathrm{T}\left ( \mathbf{C}\mathbf{C}^\mathrm{T}+\gamma  \mathbf{I} \right )^{-1},
\end{equation}
where $\mathbf{C}=\left [ \mathbf{c}_{1}, \mathbf{c}_{2}, \ldots, \mathbf{c}_{N_{S}}\right ]  \in \mathbb{R}^{N_{R}N_{G}N_{L}\times N_{S}}$ is the collected matrix including $N_{S}$ sum-pooled vectors, $\mathbf{Y}=\left [ \mathbf{y}_{1},\mathbf{y}_{2},\ldots, \mathbf{y}_{N_{S}} \right ] \in \mathbb{R}^{N_{Y}\times N_{S}}$ is the target matrix, and $\gamma$ is the regularization parameter. The output for the $s$-th sample can be determined by the index of the maximum element in $\mathbf{\hat{y}}_{s}^{\left ( i \right )}$.
\section{The analysis of the computational complexity}
\label{Sec:computational complexity}
We provide an analysis of the computational complexity of training the GDGESN in this section. Since the number of edges in each temporal graph is dynamic, we denote the number of edges for the $s$-th temporal graph at time $t$ by $E_{s}\left ( t \right )$. We define that the sparsity of $\mathbf{W}_{res}$ in each reservoir is $\varphi \in \left ( 0, 1\right ]$. The computation in the merged snapshot converter costs $\mathcal{O}\left ( \sum_{s=1}^{N_{S}}\sum_{t=1}^{N_{T}}E_{s}\left ( t \right )    \right )$. The computational complexity in each encoder is $\sum_{s=1}^{N_{S}}\sum_{t=1}^{N_{T}}\varphi N_{R}^{2}E_{s}\left ( t \right )$. The computational complexity of training the linear classifier is $\mathcal{O}\left ( \left ( N_{G}N_{L}N_{R} \right ) ^{2}\left ( N_{S}+N_{G}N_{L}N_{R} \right )  \right )$. It is obvious that the computational complexity of the proposed model in the training phase is mainly determined by the relatively larger part between the cost of running the multiple-reservoir encoder and that of the training decoding module. Therefore, the total computational complexity can be summarized as follows:
\begin{equation}
    \label{eq10}
    \max \left ( \mathcal{O}\left ( N_{G}N_{L} \varphi N_{R}^{2}\sum_{s=1}^{N_{S}}\sum_{t=1}^{N_{T}}E_{s}\left ( t \right )    \right ), \mathcal{O}\left ( \left ( N_{G} N_{L}N_{R} \right ) ^{2}\left ( N_{S}+N_{G}N_{L}N_{R} \right )  \right ) \right ).
\end{equation}
\par
In this study (see Section 5.2), $N_{G}$, $N_{L}$, and $N_{R}$ are much smaller than $N_{S}$ and  $\sum_{t=1}^{N_{T}}E_{s}\left ( t \right )$. Therefore, the computational complexity of training the GDGESN can be reduced to $\mathcal{O}\left (\sum_{s=1}^{N_{S}}\sum_{t=1}^{N_{T}} E_{s}\left ( t \right )  \right )$, which is the same with the computational complexity of DynGESN and significantly lower than many kernel-based methods~\cite{tortorella2021dynamic}.
\section{Experiments}
\label{Sec:Experiments}
\subsection{Descriptions of Datasets}
\label{Subsec:Descriptions of Datasets}
Six benchmark dissemination process classification datasets released in Ref.~\cite{oettershagen2020temporal} were used to evaluate the performances of different models. We present their details in Table.~\ref{tab:datasets}. For these six datasets, The Susceptible-Infected (SI) epidemic model~\cite{nowzari2016analysis} is used to simulate spreading processes with the corresponding infection probabilities on temporal graphs. Note that there are two categories of infections with probabilities $p_{1}$ and $p_{2}$ in every dataset, and the spreading pattern corresponding to only one probability ($p_{1}$ or $p_{2}$) exists in each temporal graph for a dataset. The datasets attached with the suffix `\_ct1' indicate that the infection probability of a spreading pattern is $p_{1}=0.5$ or $p_{2}=0.5$ in each temporal graph, and the others show that a spreading pattern with the infection probability $p_{1} = 0.2$ or $p_{2}=0.8$ exists in each temporal graph. The goal of 
the experiment is to test whether a tested model can identify two spreading patterns accurately for each dataset. In this study, we filtered empty adjacency matrices from each temporal graph sequence. 
\begin{table}[htbp]
\centering
\caption{Details of six datasets used in the experiment.}
\label{tab:datasets}
\scalebox{1}{
\begin{tabular}{lcccc}
\hline
                & \multicolumn{1}{l}{$N_{S}$} & \multicolumn{1}{l}{$N_{V}$} & \multicolumn{1}{l}{$N_{T}$} & \multicolumn{1}{l}{$\sum_{s=1}^{N_{S}}\sum_{t=1}^{N_{T}}E^{s}\left ( t \right )$} \\ \hline
                \hline
dblp\_ct1       & \multirow{2}{*}{755}        & \multirow{2}{*}{60}         & \multirow{2}{*}{48}         & \multirow{2}{*}{835714}                                                           \\
dblp\_ct2       &                             &                             &                             &                                                                                   \\ \hline
highschool\_ct1 & \multirow{2}{*}{180}        & \multirow{2}{*}{60}         & \multirow{2}{*}{205}        & \multirow{2}{*}{553013}                                                           \\
highschool\_ct2 &                             &                             &                             &                                                                                   \\ \hline
tumblr\_ct1     & \multirow{2}{*}{373}        & \multirow{2}{*}{60}         & \multirow{2}{*}{91}         & \multirow{2}{*}{1039776}                                                          \\
tumblr\_ct2     &                             &                             &                             &                                                                                   \\ \hline
\end{tabular}
}
\end{table}
\subsection{Tested models and experimental settings}
\label{Subsec:Experimental settings}
We leverage some kernel-based models in the experiments for comparison. These models can transform a temporal graph into a large-scale static graph and use kernel methods to generate final classification results. A transforming method, the directed line graph expansion (DL)~\cite{oettershagen2020temporal}, was leveraged to combine with the random walk kernel (RW)~\cite{Gartner2003Graph} and the Weisfeiler–Lehman subtree kernel (WL)~\cite{shervashidze2011weisfeiler}. These two combinations are represented by DL-RW and DL-WL, respectively. Since the transformed static graph leads to significantly high computational complexity for these two models~\cite{tortorella2021dynamic}, a simplified DL-RW method called approximate temporal graph kernel (APPR-$\mathcal{V}$)~\cite{oettershagen2020temporal}, which can sample $k$-step random
walks starting on only $\mathcal{V}$ vertices of the transformed graph, was used as another tested model. In the experiments, $\mathcal{V}$ was fixed at 250. Moreover, the prototype of GDGESN, dynamic graph echo state network (DynGESN)~\cite{micheli2022discrete} is considered as a baseline model. Note that the three kernel-based models used supported vector machine~\cite{hearst1998support} rather than the simple linear classifier leveraged by the DynGESN and the proposed GDGESN for generating classification results.
\par
\begin{table}[b]
\centering
\caption{Parameter settings of GDGESN.}
\label{tab:parameter settings}
\begin{tabular}{lll}
\hline
Parameter                               & Symbol   & Value                               \\ \hline
\hline
Spectral radius                         & $\rho$   & 0.9                                 \\
Leaking rate                            & $\alpha$ & 0.1                                 \\
Input scaling                           & $\eta$   & 1                                   \\
Density of reservoir connections        & $\varphi$& 1                                 \\
Regularization factor & $\gamma$ & 1E-3                                                             \\ 
Reservoir size                          & $N_{R}$  & 10 \\ \hline
\hline
Number of layers                        & $N_{L}$  & $\left [ 1, 2, ..., 4\right ]$     \\
Number of groups                 & $N_{G}$ & $\left [ 1,2,3 \right ]$      \\ \hline
\end{tabular}
\end{table}
For the proposed GDGESN, the parameter settings are listed in Table~\ref{tab:parameter settings}. We kept values of the spectral radius, the leaking rate, the input scaling, and the regularization factor the same as those of the DynGESN reported in Ref.~\cite{tortorella2021dynamic}. We fixed the density of the reservoir connections and the reservoir size to be 1E-3 and 10, respectively. The number of layers and the number of groups were searched in the ranges of $\left [ 1, 2, \dots, 4\right ]$, and $\left [ 1, 2, 3\right ]$, respectively. We set the size of the sliding window at $\omega^{\left ( g \right ) } = 2g-1$ for $g=1,2, \dots N_{G}$. Note that the target of this study is to show the classification improvement in performances brought about by the merged snapshot strategy in the GDGESN. Therefore we did not consider searching the key parameters of encoders and the linear classifier for extreme performances. The computational environment is an Intel (R) Core i9-7900X CPU with 96GB RAM of DDR4 2666MHz.
\par
For the partition of datasets, each dataset was evenly separated into ten parts. We cyclically picked up nine of them for the training set and the rest for the testing set for cross-validation. Based on each partition, we randomly initialized the proposed model 20 times and reported the average performances.
\subsection{Evaluation metrics}
The accuracy rate is given by the following evaluations, which can be formulated as follows:
\begin{equation}
    \mathrm{Acc} = \frac{\mathrm{The} \ \mathrm{number} \ \mathrm{of} \ \mathrm{correct} \ \mathrm{classified} \ \mathrm{temporal} \ \mathrm{graphs}}{\mathrm{The} \ \mathrm{number} \ \mathrm{of} \ \mathrm{total} \ \mathrm{temporal} \ \mathrm{graphs}} \times 100\%.
\end{equation}
\subsection{Experimental results}
\label{subsec:Experimental results}
\begin{figure}[t]
    \centering
    \includegraphics[scale=0.7]{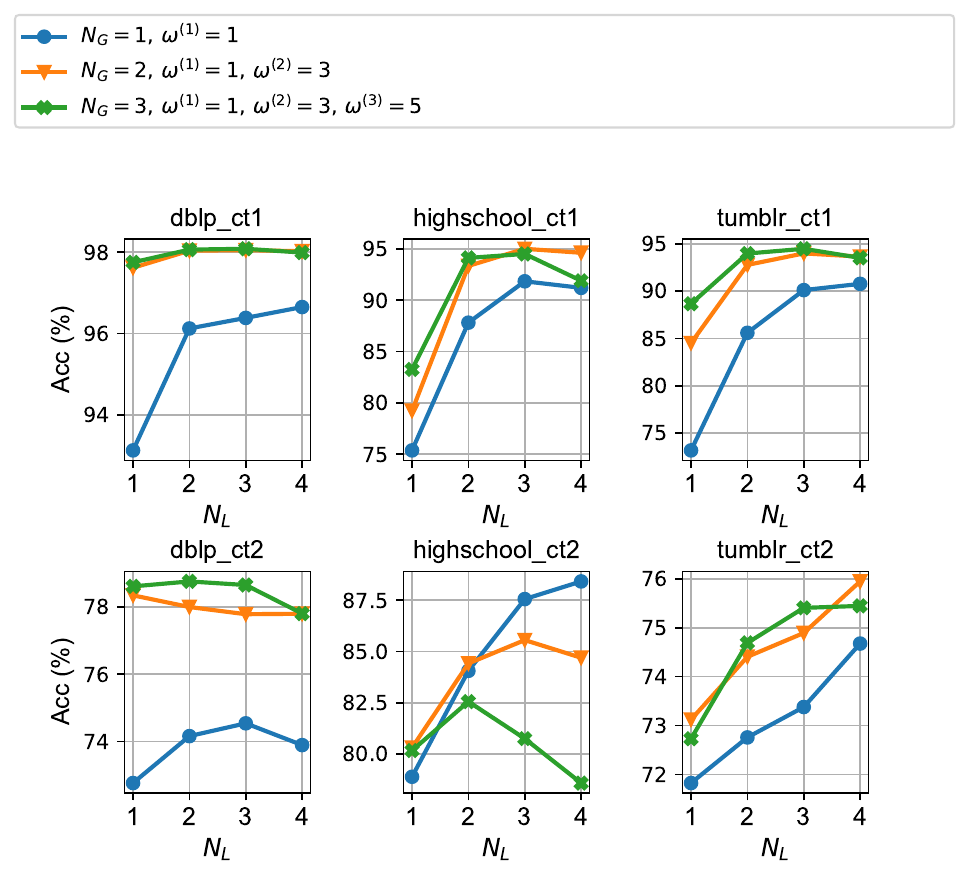}
    \caption{Average classification performances of the proposed GDGESN with variations of $N_{L}$ and $N_{G}$ on six DPC datasets}
    \label{fig:performances}
\end{figure}
To investigate the impacts brought about by the merged snapshot converter of the proposed model, we show the average classification performances of the proposed model with different combinations of $N_{L}$ and $N_{G}$ on six datasets in Fig.~\ref{fig:performances}. We notice that the GDGESN with $N_{G}>1$ outperforms the GDGESN with $N_{G}=1$ when varying $N_{L}$ from one to four on all the DPC datasets except highschool\_ct2. Specifically, the classification performances of the GDGESN with $N_{G}>1$ obviously surpass those with $N_{G}=1$ when $N_{L}=1$. These results indicate that multi-timescale spatiotemporal inputs generated by the merged snapshot converter can significantly improve the classification performances of the GDGESN for the tested DPC datasets.
\par
The best average classification performances of the GDGESN and other tested models reported in~\cite{tortorella2021dynamic} are listed in Table~\ref{tab:performance}. We highlight the corresponding best performances obtained among the APPR-250, the DynGESN, and the GDGESN in bold since these three models have significantly lower computational costs than DL-RW and DL-WL. The best performance of the GDGESN for each dataset is obtained under the best combination of $N_{P}$ and $N_{L}$ shown in Fig.~\ref{fig:performances}. We can see that the GDGESN outperforms the APPR-250 and the DynGESN on dblp\_ct1, dblp\_ct2, highschool\_ct1, and tumblr\_ct1. In particular, our model falls behind the DL-RW only on tumblr\_ct1. In addition, the GDGESN only has a few inferiorities of classification performances in comparison with the DynGESN on highschool\_ct2. Note that the dimension of each vertex embedding for the GDGESN is only 10, whereas that for the DynGESN is 16~\cite{tortorella2021dynamic}. By observing Fig.~\ref{fig:performances} and Table~\ref{tab:performance} jointly, we find that our model achieves the highest performances when $N_{L}<4$ on all the tested datasets except for highshcool\_ct2 and tumblr\_ct2, but the DynGESN obtains the best classification performances by setting $N_{L}=4$ on all datasets~\cite{tortorella2021dynamic}.
\begin{table}[htbp]
\centering
\caption{Average classification accuracy rates and standard deviations (\%) of the tested models on six datasets.}
\label{tab:performance}
\begin{tabular}{cccccc}
\hline
                & DL-RW                                    & DL-WL                                    & APPR-250       & DynGESN        & GDGESN      \\ \hline
dblp\_ct1       & 98.7$\pm$(0.1)                           & 98.5$\pm$(0.2)                           & 97.2$\pm$(0.2) & 97.7$\pm$(1.7) & \textbf{98.1$\pm$(1.7)} \\
dblp\_ct2       & 81.8$\pm$(0.9)                           & 76.5$\pm$(1.0)                           & 76.4$\pm$(0.9) & 74.3$\pm$(4.7) & \textbf{78.8$\pm$(4.2)} \\
highschool\_ct1 & 97.4$\pm$(0.7)                           & 99.2$\pm$(0.6) & 94.0$\pm$(1.3) & 94.4$\pm$(5.3) &   \textbf{95.0$\pm$(5.0)}         \\
highschool\_ct2 & 93.4$\pm$(1.0) & 89.3$\pm$(0.7)                           & 90.4$\pm$(1.8) & \textbf{92.8$\pm$(5.2)} &  88.4$\pm$(8.5)           \\
tumblr\_ct1     & 95.2$\pm$(0.6)                           & 94.2$\pm$(0.4)                           & 92.7$\pm$(0.3) & 93.3$\pm$(3.9) & \textbf{94.5$\pm$(4.0)} \\
tumblr\_ct2     & 77.2$\pm$(1.0)                           & 78.2$\pm$(1.3)                           & \textbf{78.4$\pm$(1.3)} & 76.8$\pm$(6.2) & 77.0$\pm$(6.3) \\ \hline
\end{tabular}
\end{table}
\section{Discussion}
\label{Sec:Conclusion}
We have proposed a new RC-based model for dealing with DPC tasks in this study. The proposed model can transform the original dissemination process into various multi-timescale dissemination processes and then extract the corresponding spatiotemporal features through fixed group-wise reservoir encoders. These features are decoded into the final classification results by a simple linear classifier. The simulation results show that our proposed model outperforms the DynGESN and even several kernel-based models on some benchmark DPC datasets. In addition, the analysis of computational complexity shows that our model has the same cost as DynGESN in the training process. Based on the above-mentioned contents, we can conclude that the proposed GDGESN can hold relatively high effectiveness and efficiency in dealing with DPC tasks.
\par
It is obvious that the ultimate performances are far from being reached
since we only used moderate values of key hyperparameters for multiple reservoirs in our model. We will continue exploring the optimal performances of the GDGESN on various DPC tasks in future.
\section*{Acknowledgements} This work was partly supported by JST CREST Grant Number JPMJCR19K2,
Japan (ZL, FK, GT) and JSPS KAKENHI Grant Numbers 23H03464
(GT), 20H00596 (KF), and Moonshot R\&D Grant No. JPMJMS2021(KF).
%
%
%
%
\bibliographystyle{splncs04}
\bibliography{my}

\begin{thebibliography}{10}
\providecommand{\url}[1]{\texttt{#1}}
\providecommand{\urlprefix}{URL }
\providecommand{\doi}[1]{https://doi.org/#1}

\bibitem{chen2018gc}
Chen, J., Wang, X., Xu, X.: Gc-lstm: Graph convolution embedded lstm for
  dynamic link prediction. arXiv preprint arXiv:1812.04206  (2018)

\bibitem{cho2014properties}
Cho, K., Van~Merri{\"e}nboer, B., Bahdanau, D., Bengio, Y.: On the properties
  of neural machine translation: Encoder-decoder approaches. arXiv preprint
  arXiv:1409.1259  (2014)

\bibitem{gallicchio2017deep}
Gallicchio, C., Micheli, A., Pedrelli, L.: Deep reservoir computing: A critical
  experimental analysis. Neurocomputing  \textbf{268},  87--99 (2017)

\bibitem{Gartner2003Graph}
G{\"a}rtner, T., Flach, P., Wrobel, S.: Graph kernels for chemical informatics.
  In: Proceedings of the 16th International Conference on Neural Information
  Processing Systems. pp. 505--512. MIT Press (2003)

\bibitem{guo2019attention}
Guo, S., Lin, Y., Feng, N., Song, C., Wan, H.: Attention based spatial-temporal
  graph convolutional networks for traffic flow forecasting. In: Proceedings of
  the AAAI conference on artificial intelligence. vol.~33, pp. 922--929 (2019)

\bibitem{hearst1998support}
Hearst, M.A., Dumais, S.T., Osuna, E., Platt, J., Scholkopf, B.: Support vector
  machines. IEEE Intelligent Systems and their applications  \textbf{13}(4),
  18--28 (1998)

\bibitem{hochreiter1997long}
Hochreiter, S., Schmidhuber, J.: Long short-term memory. Neural computation
  \textbf{9}(8),  1735--1780 (1997)

\bibitem{jaeger2002tutorial}
Jaeger, H.: Tutorial on training recurrent neural networks, covering bppt,
  rtrl, ekf and the" echo state network" approach  (2002)

\bibitem{kipf2016semi}
Kipf, T.N., Welling, M.: Semi-supervised classification with graph
  convolutional networks. arXiv preprint arXiv:1609.02907  (2016)

\bibitem{li2021spatial}
Li, M., Zhu, Z.: Spatial-temporal fusion graph neural networks for traffic flow
  forecasting. In: Proceedings of the AAAI conference on artificial
  intelligence. vol.~35, pp. 4189--4196 (2021)

\bibitem{li2023multi}
Li, Z., Liu, Y., Tanaka, G.: Multi-reservoir echo state networks with
  hodrick--prescott filter for nonlinear time-series prediction. Applied Soft
  Computing p. 110021 (2023)

\bibitem{li2022multi}
Li, Z., Tanaka, G.: Multi-reservoir echo state networks with sequence
  resampling for nonlinear time-series prediction. Neurocomputing
  \textbf{467},  115--129 (2022)

\bibitem{lukovsevivcius2012practical}
Luko{\v{s}}evi{\v{c}}ius, M.: A practical guide to applying echo state
  networks. In: Neural networks: Tricks of the trade, pp. 659--686. Springer
  (2012)

\bibitem{micheli2022discrete}
Micheli, A., Tortorella, D.: Discrete-time dynamic graph echo state networks.
  Neurocomputing  \textbf{496},  85--95 (2022)

\bibitem{nowzari2016analysis}
Nowzari, C., Preciado, V.M., Pappas, G.J.: Analysis and control of epidemics: A
  survey of spreading processes on complex networks. IEEE Control Systems
  Magazine  \textbf{36}(1),  26--46 (2016)

\bibitem{oettershagen2020temporal}
Oettershagen, L., Kriege, N.M., Morris, C., Mutzel, P.: Temporal graph kernels
  for classifying dissemination processes. In: Proceedings of the 2020 SIAM
  International Conference on Data Mining. pp. 496--504. SIAM (2020)

\bibitem{shervashidze2011weisfeiler}
Shervashidze, N., Schweitzer, P., Van~Leeuwen, E.J., Mehlhorn, K., Borgwardt,
  K.M.: Weisfeiler-lehman graph kernels. Journal of Machine Learning Research
  \textbf{12}(9) (2011)

\bibitem{tanaka2019recent}
Tanaka, G., Yamane, T., H{\'e}roux, J.B., Nakane, R., Kanazawa, N., Takeda, S.,
  Numata, H., Nakano, D., Hirose, A.: Recent advances in physical reservoir
  computing: A review. Neural Networks  \textbf{115},  100--123 (2019)

\bibitem{tortorella2021dynamic}
Tortorella, D., Micheli, A.: Dynamic graph echo state networks. arXiv preprint
  arXiv:2110.08565  (2021)

\bibitem{vaswani2017attention}
Vaswani, A., Shazeer, N., Parmar, N., Uszkoreit, J., Jones, L., Gomez, A.N.,
  Kaiser, {\L}., Polosukhin, I.: Attention is all you need. Advances in neural
  information processing systems  \textbf{30} (2017)

\end{thebibliography}
\end{document}